\documentclass[11pt,a4paper]{article}
\usepackage{times,latexsym}
\usepackage{url}
\usepackage[T1]{fontenc}

\usepackage{caption}
\usepackage{tikz}
\usepackage{color}
\usepackage{adjustbox}
\usepackage{booktabs}
\usepackage{makecell}
\usepackage{amsfonts}
\usepackage{amsthm}
\usepackage{amsmath}
\usepackage{amssymb}
\usepackage{bbm}
\usepackage{multirow}
\usepackage{xfrac}
\usepackage{scalerel}
\usepackage{float}
\usepackage{inconsolata} 
\usepackage[belowskip=-10pt,aboveskip=10pt]{caption}
\usepackage{subcaption}
\usepackage{enumitem}
\setlist{nolistsep}
\usepackage{algorithm, algorithmicx, algpseudocode}

\usepackage{parskip}
\usepackage[acceptedWithA]{tacl2021v1}

\usepackage[]{tacl2021v1}

\usepackage{xspace,mfirstuc,tabulary}

\newcommand{\defeq}[0]{\mathrel{\stackrel{\textnormal{\tiny def}}{=}}}

\usepackage{emoji}

\newcommand{\tree}{\emoji[twitter]{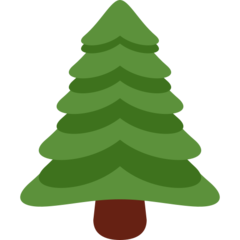}}

\usepackage[textsize=tiny]{todonotes}

\usepackage[nameinlink]{cleveref}
\crefname{section}{\S}{\S\S}
\Crefname{section}{\S}{\S\S}
\crefname{table}{Table}{}
\crefname{figure}{Fig.}{Figs.}
\crefname{algorithm}{Algorithm}{}
\crefname{equation}{Equation}{}
\crefname{inequality}{Inequality}{}
\creflabelformat{inequality}{#2{\upshape(#1)}#3} 
\crefname{appendix}{App.}{}
\crefname{thm}{Theorem}{}
\crefname{prop}{Proposition}{}
\crefname{cor}{Corollary}{}
\crefname{observation}{Observation}{}
\crefname{assumption}{Assumption}{}
\crefformat{section}{\S#2#1#3}
\crefformat{footnote}{#2\footnotemark[#1]#3}

\newcommand{\vocab}{\mathcal{V}}
\newcommand{\surp}{s}
\newcommand{\yy}{\boldsymbol{y}}

\newcommand{\uidv}{\textsc{uid}_\textit{v}(\yy)}
\newcommand{\uidl}{\textsc{uid}_\textit{lv}(\yy)}
\newcommand{\uidp}{\textsc{uid}_\textit{p}(\yy)}
\newcommand{\nuidv}{\textsc{uid}_\textit{v}}
\newcommand{\nuidl}{\textsc{uid}_\textit{lv}}
\newcommand{\nuidp}{\textsc{uid}_\textit{p}}
\newcommand{\vy}{\boldsymbol{y}}
\newcommand{\vt}{\tree}
\newcommand{\calT}{\mathcal{T}}
\newcommand{\calY}{\mathcal{Y}}
\newcommand{\yield}{\texttt{linearize}}
\newcommand{\f}{\texttt{f}}
\newcommand{\uid}{\textsc{uid}}
\newcommand{\ent}{\mathrm{H}}
\newcommand{\one}{\mathbbm{1}}
\newcommand{\lang}{\ell}
\newcommand{\plang}{p_{\lang}}
\newcommand{\parse}{\texttt{parse}}

\newcommand{\argmax}{\mathrm{argmax}}

\newcommand{\eos}{\textsc{eos}}

\newcommand{\vtheta}{\boldsymbol{\theta}}
\newcommand{\ptheta}{p_{\scaleto{\vtheta}{4pt}}}

\newcommand{\plangf}{p_{\lang_{\f}}}

\newcommand{\variantApprox}{\textsc{Approx}}
\newcommand{\variantRandom}[1]{\textsc{Random}\textsubscript{#1}}
\newcommand{\variantReal}{\textsc{Real}}
\newcommand{\variantReverse}{\textsc{Reverse}}
\newcommand{\variantOV}{\textsc{Efficient-OV}}
\newcommand{\variantVO}{\textsc{Efficient-VO}}
\newcommand{\variantDlLoc}{\textsc{Min-Dl-Loc}}
\newcommand{\variantDlOpt}{\textsc{Min-Dl-Opt}}
\newcommand{\variantSortFreq}{\textsc{Sort-Freq}}
\newcommand{\variantSortFreqRev}{\textsc{Sort-Freq-Rev}}

\newcommand{\permutefunc}{\texttt{g}}

\newif\iftaclinstructions
\taclinstructionsfalse %
\iftaclinstructions
\renewcommand{\confidential}{}
\renewcommand{\anonsubtext}{(No author info supplied here, for consistency with
TACL-submission anonymization requirements)}
\newcommand{\instr}
\fi

\iftaclpubformat %

\else

\fi

\newcommand{\ucambridge}{3}
\newcommand{\ethz}{2}
\newcommand{\MIT}{1}
\newcommand{\saar}{4}
\newcommand{\ucirvine}{5}

\title{A Cross-Linguistic Pressure for \\ Uniform Information Density in Word Order}

\author{Thomas Hikaru Clark$^{\MIT}~\;~$Clara Meister$^{\ethz}$~\;~Tiago Pimentel$^{\ucambridge}$~\;~Michael Hahn$^{\saar}$\\
\textbf{Ryan Cotterell$^{\ethz}$~\;~Richard Futrell$^{\ucirvine}$~\;~Roger Levy$^{\MIT}$}\\
  $^{\MIT}$MIT~\;~$^{\ethz}$ETH Z\"{u}rich~\;~ $^{\ucambridge}$University of Cambridge~\;~$^{\saar}$Saarland University~\;~$^{\ucirvine}$UC Irvine\\
  \texttt{\href{mailto:thclark@mit.edu}{thclark@mit.edu}}~\;~\texttt{\href{mailto:meistecl@inf.ethz.ch}{meistecl@inf.ethz.ch}}~\;~\texttt{\href{mailto:tp472@cam.ac.uk}{tp472@cam.ac.uk}}\\\texttt{\href{mailto:mhahn@lst.uni-saarland.de}{mhahn@lst.uni-saarland.de}}~\;~\texttt{\href{mailto:ryan.cotterell@inf.ethz.ch}{ryan.cotterell@inf.ethz.ch}}\\ \texttt{\href{mailto:rfutrell@uci.edu}{rfutrell@uci.edu}}~\;~ \texttt{\href{mailto:rplevy@mit.edu}{rplevy@mit.edu}}}

\date{}

\begin{document}
\maketitle

\begin{abstract}
While natural languages differ widely in both canonical word order and word order flexibility, their word orders still follow shared cross-linguistic statistical patterns, often attributed to functional pressures. In the effort to identify these pressures, prior work has compared real and counterfactual word orders. Yet one functional pressure has been overlooked in such investigations: the uniform information density (UID) hypothesis, which holds that information should be spread evenly throughout an utterance. Here, we ask whether a pressure for UID may have influenced word order patterns cross-linguistically. To this end, we use computational models to test whether real orders lead to greater information uniformity than counterfactual orders. In our empirical study of 10 typologically diverse languages, we find that: (i) among SVO languages, real word orders consistently have greater uniformity than reverse word orders, and (ii) only linguistically implausible counterfactual orders consistently exceed the uniformity of real orders. These findings are compatible with a pressure for information uniformity in the development and usage of natural languages.\footnote{Code for reproducing our experiments is available at \url{https://github.com/thomashikaru/word-order-uid}.\label{note1}}

\end{abstract}

\vspace{-6pt}
\section{Introduction}
\vspace{-2pt}

Human languages differ widely in many respects, yet there are patterns that appear to hold consistently across languages.
Identifying explanations for these patterns is a fundamental goal of linguistic typology.
Furthermore, such explanations may shed light on the cognitive pressures underlying and shaping human communication.\looseness=-1 

This work studies the \textit{uniform information density}
(UID) hypothesis as an explanatory principle for word order patterns \cite{fenk1980konstanz, genzel2002entropy, AylettTurk2004,jaeger2010redundancy,meister2021revisiting}.
The UID hypothesis posits a communicative pressure to avoid spikes in information within an utterance, thereby keeping the information profile of an utterance relatively close to uniform over time. 
While the UID hypothesis has been proposed as an explanatory principle for many linguistic phenomena, e.g., speakers' choices when faced with lexical and syntactic alternations
\citep{Levy_Jaeger_2006}, its relationship to word order patterns has received limited attention, with the notable exception of \citet{maurits2010why} and \citet{jain-etal-2018-uniform}.\looseness=-1 

Our work investigates the relationship between UID and word order patterns, differing from prior work in several ways. We (i) use Transformer language models (LMs) \cite{vaswani2017attention} to estimate information-theoretic operationalizations of information uniformity; (ii) analyze large-scale naturalistic datasets of 10 typologically diverse languages; and (iii) compare a range of theoretically motivated counterfactual grammar variants.
\looseness=-1

Experimentally, we find that among SVO languages, the real word order has a more uniform information density than nearly all counterfactual word orders; the only orders that consistently exceed real orders in uniformity are generated using an implausibly strong bias for uniformity, at the cost of expressivity. 
Further, we find that counterfactual word orders that place verbs before objects are more uniform than ones that place objects before verbs in nearly every language.\looseness=-1

Our findings suggest that a tendency for uniform information density may exist in human language, with two potential sources:
(i) word order rules, with SVO order generally being more uniform than SOV;
and (ii) choices made by speakers, who use the flexibility present in real languages to structure information more uniformly at a global level (and not only in a small number of isolated constructions). 
\looseness=-1

\section{Functional Pressures in Language}

\subsection{Linguistic Optimizations}
A number of linguistic theories link cross-linguistic patterns to functional pressures. 
For example, both the grammatical rules of a language and speakers' choices (within the space of grammatically acceptable utterances) are posited to reflect a trade-off between effort and robustness: 
shorter and simpler structures are easier to produce and comprehend, but longer and more complex utterances can encode more information \citep{gabelentz1901sprachwissenschaft, zipf1935psychobiology,hawkins1994performance,hawkins2004efficiency,hawkins2014crosslinguistic,haspelmath2008parametric}.
Another such functional pressure follows from the principle of dependency length minimization (DLM), which holds that, in order to minimize working memory load during comprehension, word orders should place words in direct dependency relations close to each other  \citep{rijkhoff1986word,rijkhoff1990explaining,hawkins1990parsing,hawkins1994performance,hawkins2004efficiency,hawkins2014crosslinguistic,grodner2005consequences,Gibson_1998,gibson2000dependency,bartek2011search,temperley2018minimizing,futrell2020dependency}.
A growing body of work has turned to information theory, the mathematical theory of communication \citep{shannon1948mathematical}, to formalize principles that explain linguistic phenomena \citep{jaeger2011language,GIBSON2019389,pimentel2021nonoptimal}.  One such principle is that of uniform information density.\looseness=-1

\subsection{Uniform Information Density}

According to the uniform information density (UID) hypothesis, speakers tend to spread information evenly throughout an utterance; large fluctuations in the per-unit information content of an utterance can impede communication by increasing the processing load on the listener. 
Speakers may modulate the information profile of an utterance by selectively producing linguistic units such as optional complementizers in English \citep{Levy_Jaeger_2006,jaeger2010redundancy}. A pressure for UID in speaker choices has also been studied in specific constructions in other languages, though with mixed conclusions  \citep{zhan2018comparing, clark-etal:2022-evidence-for-availability}.\looseness=-1

Formally, the information conveyed by a linguistic signal $\vy$, e.g., an utterance or piece of text, is quantified in terms of its surprisal $\surp(\cdot)$, which is defined as $\vy$'s negative log-probability:
$\surp(\vy) \defeq - \log \plang(\vy)$. Here, $\plang$ is the underlying probability distribution over sentences $\vy$ for a language $\lang$. Note that we do not have access to the true distribution $\plang$, and typically rely on a language model with learned parameters $\vtheta$ to estimate surprisal values with a second distribution $\ptheta$.

Surprisal can be additively decomposed over the units that comprise a signal. Explicitly, for a signal $\vy$ that can be expressed as a series of linguistic units $\langle y_1, \dots, y_N\rangle$, where $y_n \in \vocab$ and $\vocab$ is a set vocabulary of words or morphemes, the surprisal of a unit $y_n$ is its negative log-probability given prior context: $s(y_n) =- \log \plang(y_n \mid \vy_{< n})$.
Note that the distribution $\plang(\cdot \mid \vy_{< n})$ has support $\overline{\vocab} \defeq\vocab \cup \{\eos\}$, where $\eos$ is a designated symbol indicating the end of a sequence;\footnote{ This symbol allows for the global normalization of $\plang$, i.e., a valid probability distribution over finite-length sequences $\vocab^*$ \cite[see][for a discussion]{Du_Hennigen_Pimentel_Meister_Eisner_Cotterell_2022}. }
a valid, complete signal $\vy= \langle y_1, \dots, y_N\rangle$ has $y_N=\eos$. 
The quantity $\surp(\vy)$ can thus likewise be expressed as $\surp(\vy) = \sum_{n=1}^N s(y_n)$. 
Assuming that we have a fixed amount of information to convey and that high-surprisal items are disproportionately difficult to process,\footnote{
Most empirical results \cite{hale2001probabilistic,levy2008expectation, shain_meister_pimentel_cotterell_levy_2022} suggest that a word's processing effort is directly proportional to its surprisal. Yet there is also evidence of a superlinear relationship, which would imply a preference by the comprehender for UID \cite{meister2021revisiting, hoover_2022}.\looseness=-1} it can be shown mathematically that spreading information evenly throughout a signal optimizes ease of processing for the comprehender 
\citep{Levy_Jaeger_2006,smith2013effect,levy2018communicative,meister2021revisiting}.\looseness=-1

While the UID hypothesis is often discussed in the context of speaker choices, it has also been presented as a general cognitive constraint that might influence reading times \citep{meister2021revisiting}, speech duration \citep{Pimentel2021_surprisal_duration}, and word lengths \citep{piantadosi2011word}. 
Selection for UID has also been discussed as a potential evolutionary pressure on language that can explain typological differences \cite{jaeger2011language}. 
Within this literature, there is not a consensus on how to formally operationalize UID. 
For example, \citet{frank2008speaking} measure regression of surprisal towards a language-wide mean; \citet{collins2014} and \citet{bloem2016testing} consider more local changes in surprisal in their quantification of UID.\looseness=-1

In this work, we consider three metrics for operationalizing UID \citep{meister2021revisiting}:
\begin{align}
&\uidv \defeq \frac{1}{N}\sum_{n=1}^N (s(y_{n}) - \mu )^2  \label{eq:uid-v}
\end{align}
In \cref{eq:uid-v}, $\nuidv$ is the mean within-sentence variance of word surprisals, where $\mu = \frac{1}{N}\sum_{n=1}^{N} s(y_n)$ is a sentence-level mean.
\begin{align}
&\uidl \defeq \frac{1}{N\!-\!1}\sum_{n=2}^N (s(y_{n}) - s(y_{n-1}) )^2 \label{eq:uid-l}
\end{align}
In \cref{eq:uid-l}, $\nuidl$ quantifies the average word-to-word change in surprisal, a more localized measure \cite{collins2014}. Intuitively, this is maximized when high-surprisal words alternate with low-surprisal words, and minimized when words appear in sorted order by information content.  
\begin{align}
&\uidp \defeq \frac{1}{N}\sum_{n=1}^N s(y_{n})^k \label{eq:uid-p}
\end{align}
In \cref{eq:uid-p}, $\nuidp$ is a power mean with $k>1$, which disproportionately increases in the presence of larger surprisal values.\footnote{This metric suggests a super-linear processing cost for surprisal.\looseness=-1} 
Note that for all of these operationalizations, lower values correspond to greater uniformity.\footnote{We note that, while a fully uniform language would have value 0 for $\nuidv$ and $\nuidl$, it would not for $\uidp$, so the metrics are not directly comparable. 
}\looseness=-1

\section{Counterfactual Language Paradigm}
\begin{figure*}
        \centering
        \includegraphics[width=0.95\linewidth]{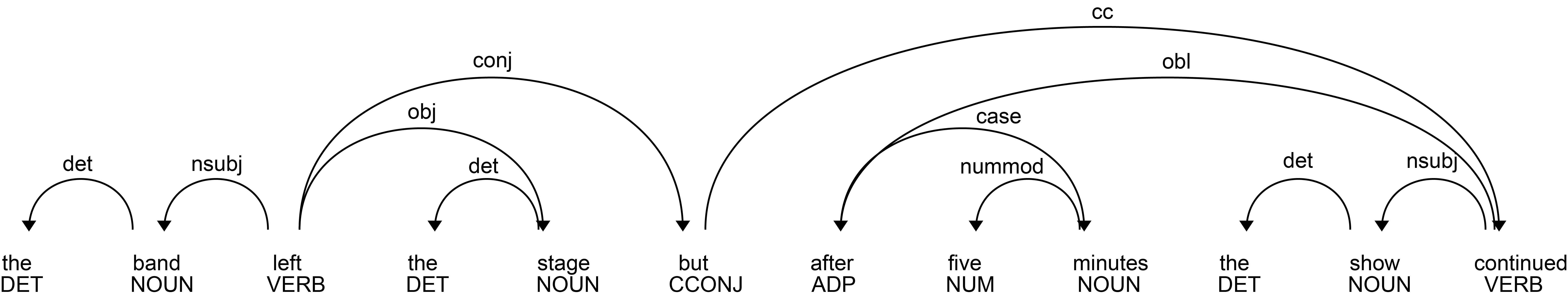}
        \caption{An example dependency tree showing syntactic relationships according UD, transformed so that function words are heads (\cref{sec:counterfactual-grammars}). Arrows point from heads to dependents. \looseness=-1}
        \label{fig:example-tree}
\end{figure*}
Following prior work that has used counterfactual languages to study the functional pressures at play in word order patterns, we investigate to what degree a language's word order shows signs of optimization for UID. 
In this approach, a corpus of natural language is compared against a counterfactual corpus containing minimally changed versions of the same sentences, where the changes target an attribute of interest, e.g., the language's word order.
For example, several studies of DLM have compared syntactic dependency lengths in real and counterfactual corpora, generated by permuting the sentences' word order either randomly \citep{ferrericancho2004euclidean,liu2008dependency} or deterministically by applying a counterfactual grammar \citep{gildea2010grammars, Gildea_Jaeger_2015, Futrell_Mahowald_Gibson_2015, futrell2020dependency}. 
Similarly, we will compare measures of UID in real and counterfactual corpora to investigate whether real languages' word orders exhibit more uniform information density than alternative realizations.\looseness=-1 

\subsection{Formal Definition}\label{sec:counterfactual-grammars-def}
We build on the counterfactual generation procedure introduced by \citet{Hahn_Jurafsky_Futrell_2020} to create parallel corpora.
This procedure operates on sentences' dependency parses. 
Formally, a dependency parse $\vt$ of a sentence $\vy$ is a directed tree with one node for every word, where each word in $\vy$, with the exception of a designated root word, is the child of its (unique) syntactic head; see \citet{Zmigrod_Vieira_Cotterell_2020} for a discussion of the role of the root constraint in dependency tree annotation. 
Each edge in the tree is annotated with the syntactic relationship between the words connected by that edge; see \cref{fig:example-tree} for an example.
Here we use the set of dependency relations defined by the Universal Dependencies (UD) paradigm \cite{de_Marneffe_Manning_Nivre_Zeman_2021}, though we follow \citet{Hahn_Jurafsky_Futrell_2020} in transforming dependency trees such that function words are treated as heads, leading to representations closer to those of standard syntactic theories; see also \citet{DBLP:conf/acludw/GerdesGKP18}.

\begin{figure}
    \centering
    \includegraphics[width=\linewidth]{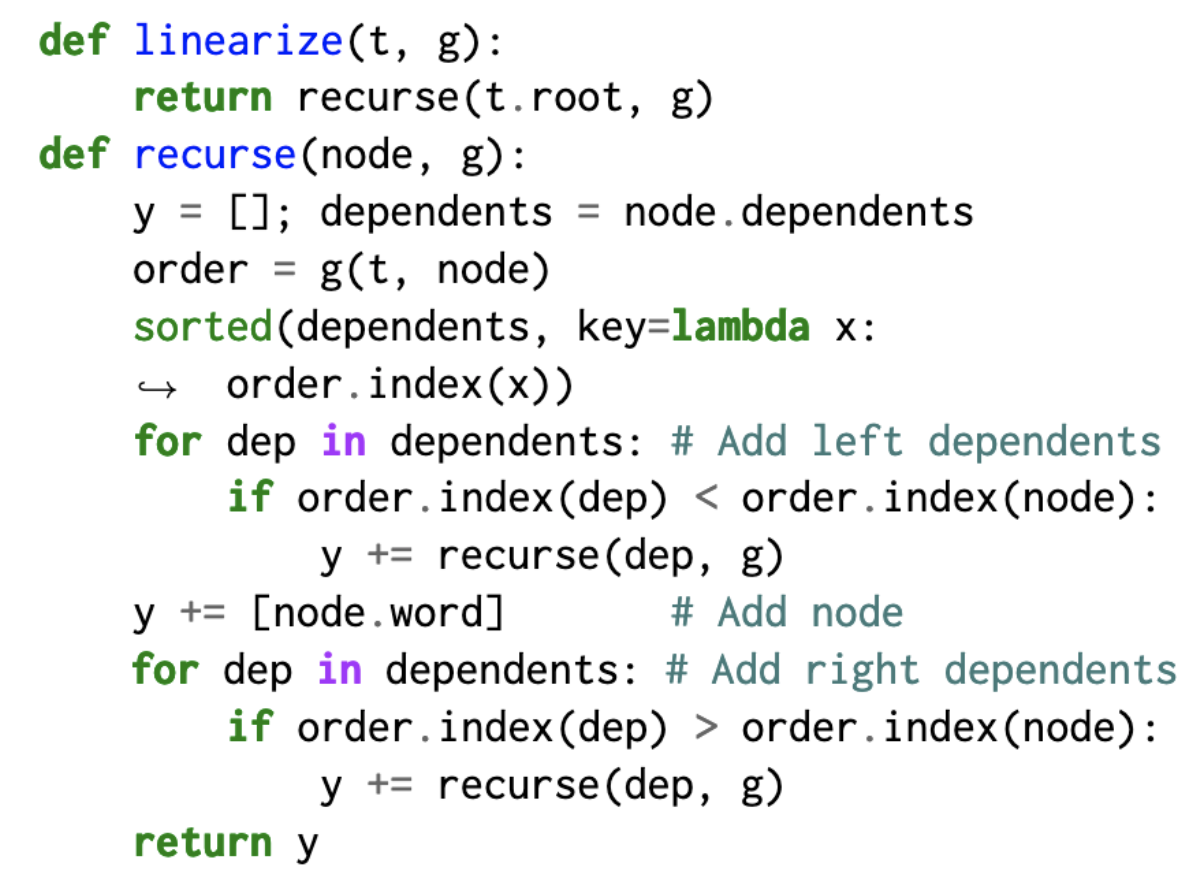}
    \vspace{-4pt}
    \caption{Pseudo-code to linearize a dependency tree $\vt$ according to a grammar's ordering function $\permutefunc$. In this code, each \texttt{node} contains a \texttt{word} and its syntactic \texttt{dependents}.}
    \label{fig:yield_algorithm}
\end{figure}

\paragraph{Tree linearization.} While syntactic relationships are naturally described hierarchically, sentences are produced and processed as linear strings of words.
Importantly, there are many ways to linearize a dependency parse $\vt$'s nodes into a string $\vy$. 
Concretely, a grammar under our formalism is defined by an 
\textit{ordering function} \cite[see][]{Kuhlmann_2010}
$\permutefunc(\cdot,\cdot)$ which takes as arguments a dependency parse and a specific node in it, and returns an ordering of the node and its dependents.
For each node, its dependents are arranged from left to right according to this ordering; any node without dependents is trivially an ordered set on its own. 
This process proceeds recursively to arrive at a final ordering of all nodes in a dependency tree, yielding the final string $\vy$. 
Pseudo-code for the linearization of a tree $\vt$ based on an ordering function $\permutefunc$ is given in \cref{fig:yield_algorithm}.

\paragraph{Simplifying assumptions.} 
One consequence of this formalism is that all counterfactual orders correspond to projective trees, i.e., trees with no crossing dependencies. While projectivity is a well-attested cross-linguistic tendency, human languages do not obey it absolutely \cite{ferrericancho2018crossing,  yadav2021dependency}. 
Within the space of projective word order interventions allowed by this formalism, the grammars which we borrow from \citet{Hahn_Jurafsky_Futrell_2020} enforce two additional simplifying constraints. First, the relative positioning (left or right) between the head and dependent of a particular relation is fixed. Second, the relative ordering of different relations on the same side of a head is also fixed. We denote grammars which satisfy both constraints as \textit{consistent}.
Notably, natural languages violate both of these assumptions to varying degrees.
For example, even in English -- a language with relatively strict word order -- adverbs can generally appear before or after their head.
While these simplifications mean that the formalism cannot perfectly describe natural languages, it provides a computationally well-defined method for intervening on many features of word order. In particular, the consistent grammars of \citet{Hahn_Jurafsky_Futrell_2020} are parameterized by a set of scalar weights corresponding to each possible syntactic relation; the ordering function thus reduces to sorting each head's dependents based on their weight values. Notably, \citet{Hahn_Jurafsky_Futrell_2020} also introduced a method for optimizing these grammars for various objective functions by performing stochastic gradient descent on a probabilistic relaxation of the grammar formalism; we use several of these grammars (described in \cref{sec:counterfactual-grammars}) in our subsequent analysis.

\paragraph{Creating counterfactual word orderings.}
The above paradigm equips us with the tools necessary for  systematically altering sentences' word orderings, which in turn, enables us to create counterfactual corpora. 
Notably, the large corpora we use in this study contain sentences as strings, not as their dependency parses. We therefore define our counterfactual grammar intervention as the output of a (deterministic) word re-ordering function $\f : \calY \rightarrow \calY$, 
where $\calY \defeq \vocab^*$ is the set of all possible sentences that can be constructed using a language's vocabulary $\vocab$.\footnote{For notational brevity, we leave the dependency of $\vocab$ on $\lang$ implicit as it should be clear from context.} 
This function takes as input a sentence from our original language and outputs a sentence with the counterfactual word order defined by a given ordering function $\permutefunc$. 
We decompose this function into two steps:
\begin{align}
    \f(\vy) = \yield( \parse(\vy), \permutefunc)
\end{align}
We use a state-of-the-art parser \cite{udpipe:2017} to implement $\parse : \calY \rightarrow\calT$ where $\calT$ is the set of all dependency parses.
Specifically, we define $\parse(\vy) = \argmax_{\vt \in \mathcal{T}} \,\, p(\vt \mid \vy)$ 
for a learned conditional probability distribution over possible parses $p(\cdot \mid \vy)$.
We then obtain the linearized form of the resulting tree by supplying it and the ordering function $\permutefunc$ to $\yield$
, as defined above. 
Collectively, the outputs of this process (parallel datasets differing only in word order) are referred to as \textit{variants}.
Importantly, $\f$ here is a deterministic function; one could instead consider $\f$ to be probabilistic in nature, with each sentence $\vy$ having a distribution over tree structures $\vt$. 
We discuss the implications of this choice in \cref{sec:limitations}.\looseness=-1

\subsection{Counterfactual Grammar Specifications}\label{sec:counterfactual-grammars}

In addition to the original \variantReal{} word order, we explore the following theoretically motivated counterfactual grammars for each language.\looseness=-1

\paragraph{Consistent approximation to real order.} \variantApprox{} is a consistent approximation to the real word order within our formalism; it uses an ordering function parameterized by weights
that were fitted to maximize the likelihood of observed word orders for each language, as reported by \citet{Hahn_Jurafsky_Futrell_2020}.
This variant captures most of the word order features of a real language while allowing for a fair comparison to deterministic counterfactual grammars that do not model the flexibility of real language. From the perspective of the UID hypothesis, we expect this variant to be less uniform that \variantReal{} because it has less flexibility to accommodate speakers' choices that optimize for UID. \looseness=-1

\paragraph{Consistent random grammars.} We include variants 
\variantRandom{1} through \variantRandom{5}, which use ordering functions parameterized by randomly assigned weights.
This means that for a given random grammar, each dependency relation has a fixed direction (left or right), but that the directions of these relations lack the correlations observed in natural language \cite{greenberg}. Random grammars with the same numerical index share weights across languages. 

\paragraph{Consistent grammars optimized for efficiency.} 
We include two consistent grammars that are optimized for the joint objective of parseability (how much information an utterance provides about its underlying syntactic structure) and sentence-internal 
predictability, as reported by \citet{Hahn_Jurafsky_Futrell_2020}, one with OV order (\variantOV{}) and one with VO order (\variantVO{}).
For example, the \variantOV{} grammar for English would give a plausible version of a consistent and efficient grammar in the counterfactual world where English has verbs after objects.  

\paragraph{Grammars optimized for dependency length minimization.} From the same work we also take consistent grammars that are optimized for DLM, denoted as \variantDlOpt{}. While linearizations produced by these grammars are not guaranteed to minimize dependency length for any particular sentence, they minimize the expected average dependency length of a large sample of sentences in a language. In addition, we include \variantDlLoc{}, an inconsistent grammar that applies the projective dependency-length minimization algorithm of \citet{Gildea_Temperley_2007} at the sentence level, leading to sentences with minimal DL but without the constraint of consistency.
\paragraph{Frequency-sorted grammars.} \variantSortFreq{} is an inconsistent grammar which orders words in a sentence from highest to lowest frequency, ignoring dependency structure altogether. We use this ordering as a heuristic baseline for which we expect UID to hold relatively strongly: low-frequency elements, which tend to have higher surprisal even if solely from their less frequent usage \citep{ellis2002frequency}, are given more context, and thus should have smaller surprisals than if they occurred early; more conditioning context tends to reduce the surprisal of the next word \cite{luke2016limits}. 
We also test \variantSortFreqRev{}, ordering words from least to most frequent, which for analogous reasons we expect to perform poorly in terms of UID. However, both of these orderings lead to massive syntactic ambiguity by introducing many string collisions -- any two sentences containing the same words in different orders would be linearized identically. This eliminates word order as a mechanism for expressing distinctions in meaning, so these orders are implausible as alternatives to natural languages \cite{mahowald2022experimentally}. 
\paragraph{Reverse grammar.} Finally, we also include the \variantReverse{} variant, where the words in each sentence appear in the reverse order of the original. 
This variant preserves all pairwise distances between words within sentences and has identical dependency lengths as the original order, thus isolating the effect of linear order on information density from other potential influences. Notably, if the original language happens to be perfectly consistent, then \variantReverse{} will also satisfy consistency; in practice, this is unlikely to hold with natural languages.\looseness=-1

\subsection{UID and Counterfactual Grammars} \label{sec:analyzing-uid}

\begin{figure*}
        \centering
        \includegraphics[width=0.9\linewidth,trim={0 5cm 0 0},clip]{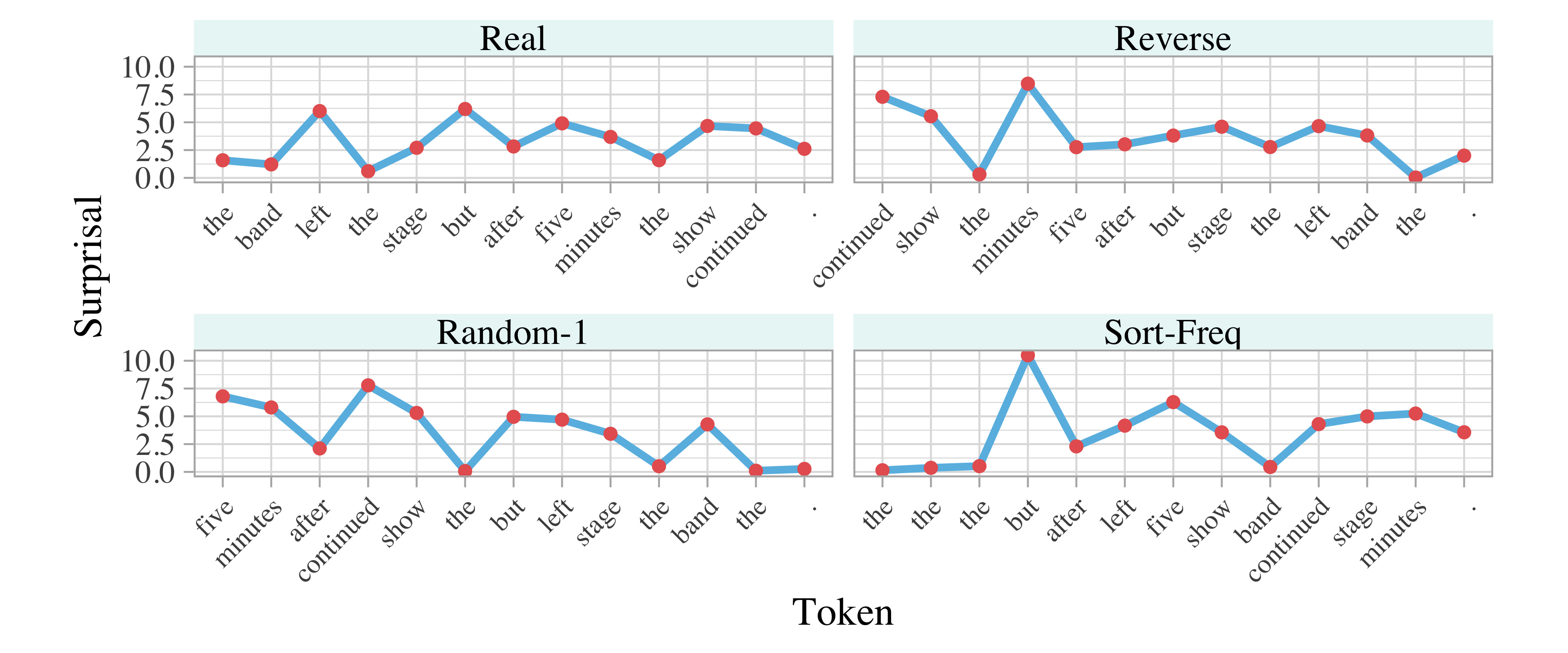}
        \vspace{-5pt}
        \caption{The same source sentence according to 4 real and counterfactual orderings. \looseness=-1}
        \label{fig:example-sentence}
\end{figure*}

Let $\plang(\vy)$ be the probability distribution over sentences $\vy$ for a language of interest $\lang$. 
We can define a language's UID score as the expected value of its sentences' UID scores, where we overload the $\uid$ function to take either a sentence $\vy$ or an entire language $\lang$:
\begin{align}
    \uid(\lang) &\defeq \sum_{\vy \in \calY} \plang(\vy)\,\uid(\vy)
\end{align}
where sentence-level $\uid$ can be $\uidv$, $\uidl$, or $\uidp$. In practice, we estimate this language-level UID score using a Monte-Carlo estimator, taking the mean sentence-level UID score across a held-out test set $S_{\lang}$ of sentences $\vy$ in language $\lang$, where we assume $\vy\sim\plang$:
\begin{equation}
\widehat{\uid}(\lang) \defeq \frac{1}{|S_{\lang}|} \sum_{\vy \in S_{\lang}} \uid(\vy)
\end{equation}
Similarly, the expected surprisal (or Shannon entropy, $\ent$) of this language is computed as:\looseness=-1
\begin{align}
    \ent(\lang) &\defeq - \sum_{\vy \in \calY} \plang(\vy)\,\log 
    \plang(\vy)\label{eq:uidlang}
\end{align}

We evaluate how well a language model $\ptheta$ approximates $\plang$ by its cross-entropy:
\begin{align}
    \ent(\plang, \ptheta) = - \sum_{\vy \in \calY} \plang(\vy)\,\log \ptheta(\vy)
\end{align} 
where a smaller value of $\ent$ implies a better model. Again using a Monte Carlo estimator, we measure cross-entropy using the held-out test set $S_{\lang}$: 
\begin{align}
    \widehat\ent(\plang, \ptheta) = - \frac{1}{|S_{\lang}|} \sum_{\vy \in S_{\lang}} \log \ptheta(\vy)
\end{align}
This is simply the \emph{mean surprisal} that the model assigns to a corpus of naturalistic data.

These computations can also be applied to counterfactual variants of a language. 
Let $\lang_\f$ stand for a language identical to $\lang$, but where its strings have been transformed by $\f$;
this language's distribution over sentences would be $\plangf(\vy) = \sum_{\vy' \in \calY} \plang(\vy') \one\{\vy = \f(\vy')\}$. 
Since entropy is non-increasing over function transformations (by Jensen's inequality), it follows that:
\begin{align} \label[inequality]{ineq:ent_injection}
    \ent(\lang) \geq \ent(\lang_\f)
\end{align}
Further, if our counterfactual generation function $\f$ is a bijection 
-- meaning that each input string gets mapped to a distinct output string and each output string has an input that maps to it -- then we can create a second function
$\f^{-1}: \calY  \rightarrow \calY$, which would generate $\lang$ from $\lang_\f$.
Then, the following holds:
\begin{align} \label{eq:ent_bijection}
    \ent(\lang) \geq \ent(\lang_\f) \geq \ent(\lang_{\f^{-1} \circ\, \f}) = \ent(\lang)
\end{align}
i.e., it must be that $\ent(\lang) = \ent(\lang_\f)$.
Reversing a sentence is an example of a bijective function, and thus \cref{eq:ent_bijection} holds necessarily for the pair of \variantReal{} and \variantReverse{} variants; the counterfactual generation procedure thus should not produce differences in mean surprisal between these variants. 
At the same time, bijectivity does not necessarily hold for our other counterfactual transformations and is violated to a large degree when mapping to \variantSortFreq{} and \variantSortFreqRev{}. Thus in general, we can only guarantee \cref{ineq:ent_injection}. 

\looseness=-1

Crucially, however, the transformation $\f$ might change the UID score of such a language, allowing us to evaluate the impact of word order on information uniformity. 
As a simple example, consider the language $\lang_1$ that places a uniform distribution over only four strings: $aw$, $ax$, $by$, and $bz$.
In this language, the first and second symbols always have 1 bit of surprisal, and the end of the string has 0 bits of surprisal.
If the counterfactual language $\lang_2$ is the reverse of $\lang_1$, we have a uniform distribution over the strings $wa$, $xa$, $yb$, and $zb$. Here, the first symbol always has 2 bits of surprisal, and the second symbol and end of sentence always have zero bits, as their values are deterministic for a given initial symbol. While the mean surprisal per symbol is the same for $\lang_1$ and $\lang_2$,  $\lang_1$ has more uniform information density than $\lang_2$.

\section{Limitations} \label{sec:limitations}

\subsection{Use of Counterfactual Grammars} \label{sec:use-of-cf-grammars}

\paragraph{Real word orders are not consistent.}
The consistent grammars borrowed from \citet{Hahn_Jurafsky_Futrell_2020} assume that the direction of each syntactic relation, as well as the relative ordering of dependents on the same side of a head, are fixed. This is not generally true of natural languages.
We address this difference by including the variant \variantApprox{} as a comparison to the counterfactual variants, which are constrained by consistency, and by including \variantReverse{} as a comparison to \variantReal{}, both of which are not constrained by consistency. 
\paragraph{Automatic parsing errors.}
Another issue is that the dependency parses extracted for each original sentence as part of the counterfactual generation pipeline may contain parsing errors. These errors may introduce noise into the counterfactual datasets that is not present in the original sentences, and may cause deviations from the characteristics that we assume our counterfactual grammars should induce. For example, \variantDlLoc{} only produces sentences with minimized dependency length if the automatic parse is correct.  

\paragraph{Deterministic parsing.}
Finally, our counterfactual generation procedure assumes a deterministic mapping from sentences to dependency trees as one of its steps. However, multiple valid parses of sentences are possible in the presence of syntactic ambiguity.
In such cases, we always select the most likely structure according to the parser, which learns these probabilities based on its training data.
Therefore, this design choice could lead to underrepresentation of certain syntactic structures when applying a transformation.
However, we note that the variants \variantReal{}, \variantReverse{}, \variantSortFreq{}, and \variantSortFreqRev{} do not depend on dependency parses and so are unaffected by this design choice.\looseness=-1

\subsection{Choice of Dataset}\label{sec:dataset-choice}
Properties of language can vary across genres and domains. When drawing conclusions about human language in general, no single dataset will be completely representative. Due to the amount of data required to train LMs, we use written corpora in this work, and use the term \textit{speaker} loosely to refer to any language producer regardless of modality.
To address potential concerns about the choice of dataset in this study, we conducted a supplementary analysis on a subset of languages using a different web corpus, which we report in \cref{sec:dataset-choice}. 

\subsection{Errors and Inductive Biases}\label{lm-inductive-biases}
 
\newcommand{\calN}{\mathcal{N}}
\newcommand{\uniform}{\mathtt{uniform}}

\paragraph{Model Errors.} 
Language model quality could impact  the estimated values of our UID metrics $\nuidv$, $\nuidp$, and $\nuidl$. 
To see why, 
consider a model $\ptheta$ that -- rather than providing unbiased estimates of $\plang$ -- is a smoothed interpolation between $\plang$ and the uniform distribution:\looseness=-1
\begin{align}
    \ptheta(y_n\mid \vy_{<n}) = \lambda\, \plang(y_n\mid \vy_{<n}) + \frac{1 - \lambda}{|\overline{\vocab}|}
\end{align}
for $\lambda \in [0,1]$.
Here, an increase in $1\!-\!\lambda$ would lead to an increase in $\ent(\plang, \ptheta)$, since the cross-entropy is only minimized when $\ptheta(\cdot \mid \vy_{<n}) = \plang(\cdot\mid \vy_{<n})$. 
This change, however, would be reflected as an \emph{increase} in uniformity, e.g., a decrease in  $\nuidv$: surprisals would be closer to uniform for smaller values of $\lambda$. Alternatively, consider the situation where a language $\lang$ has perfect information uniformity, i.e., where $\nuidv$, $\nuidp$, and $\nuidl$ are their minimum possible values. The  interpolation of  $\plang$ with any non-uniform distribution should instead \emph{decrease} the measured uniformity, at least with respect to  $\nuidv$ and  $\nuidl$.

In summary, our UID metrics could be biased either positively or negatively by the quality of our models. 
However, since our analysis focuses on the comparison of UID metrics between word order variants rather than their absolute value, this bias should not be a major concern. We use the same model architecture for all language--variant combinations, and so a bias in the UID metric corresponding to one combination should likewise be reflected in all of the metrics that it is compared to. 
Further, our results hold even when controlling for mean surprisal, as described in \cref{sec:results}.\looseness=-1

\paragraph{Inductive Biases.} Because modern LMs have been developed to model natural language, they may contain subtle biases towards the properties of real word orders or of highly resourced languages. 
Based on \cref{ineq:ent_injection}, if two probabilistic models $m_{\lang}$ and $m_{\lang_{\f}}$ were to perfectly learn the true and counterfactual distributions $\plang$ and $\plangf$, respectively, then $m_{\lang}$ should assign approximately the same or higher mean surprisal to a corpus $\{\vy^{(m)}\}_{m=1}^{M}$ from $\lang$ than $m_{\lang_{\f}}$ assigns to the counterfactual corpus from $\lang_\f$.
This implies that previous results of \citet{Gildea_Jaeger_2015}, \citet{Ravfogel_Goldberg_Linzen_2019}, \citet{Hahn_Jurafsky_Futrell_2020} and \citet{white-cotterell-2021-examining}, which found that real corpora tend to have lower average per-word surprisal than 
deterministically generated counterfactual versions of the same corpora, were in fact due to the inductive bias of the learning algorithms used to estimate surprisals.
There is a clear reason why the trigram model of \citet{Gildea_Jaeger_2015} would yield higher mean surprisals for counterfactual corpora: the transformation functions $\f$ tended to increase dependency lengths, and words in a dependent--head relation tend to have higher mutual information than other pairs of words \citep{futrell2017,futrell2019syntactic,futrell2020dependency}. Hence the transformations tended to push words that are predictive of each other outside of the conditioning window of the model \citep[see also][for similar effects]{hahn2022crosslinguistic}. The Transformer architecture we use in this work could thus also contain biases favoring features of real language, which we attempt to control for (see \cref{sec:results}).

\begin{figure*}
        \centering
        \includegraphics[width=0.9\linewidth]{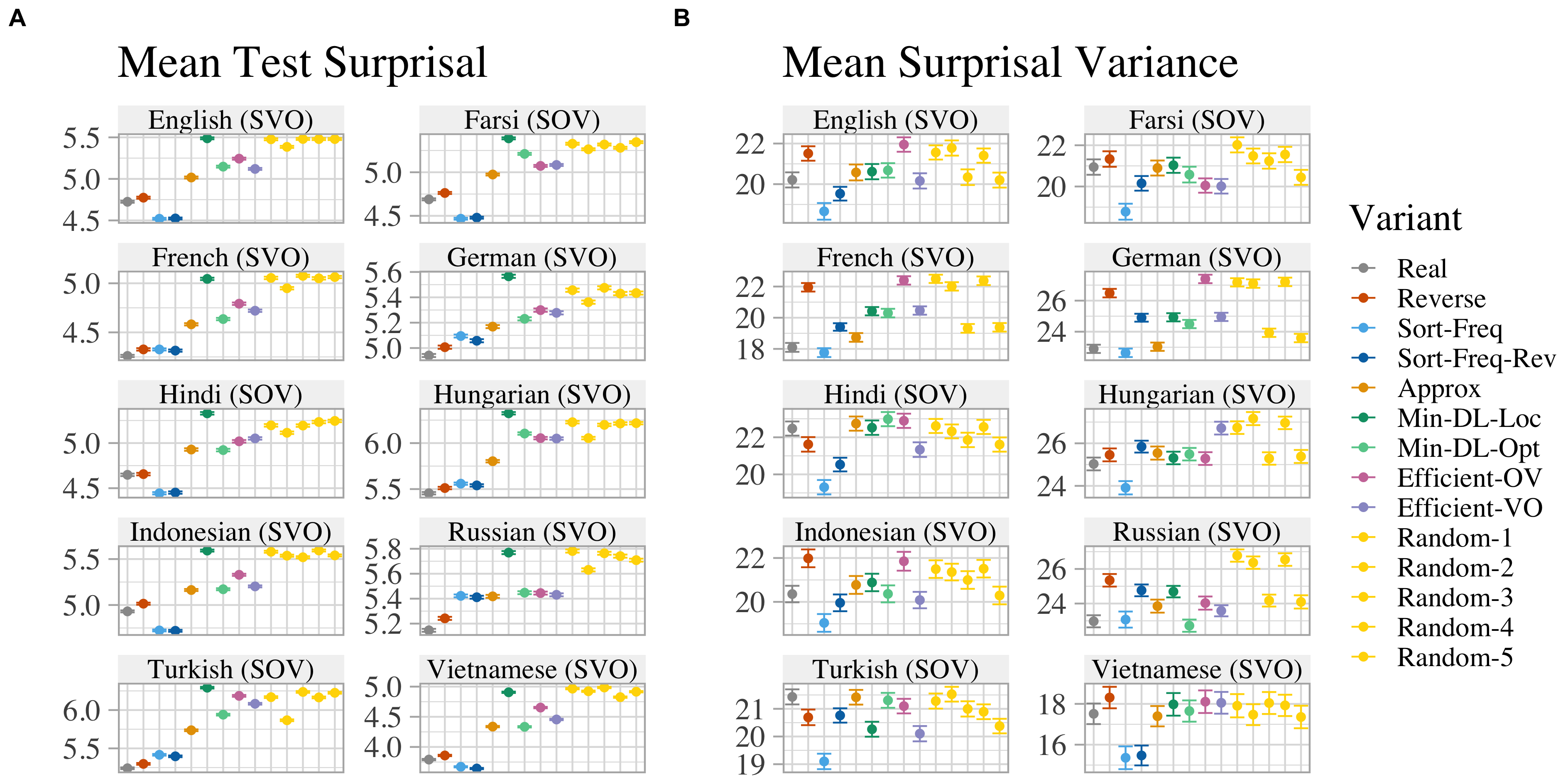}
        \caption{Mean test-set surprisal and surprisal variance of language models across real and counterfactual grammars in 10 languages. Error bars denote the 95\% CI of the mean. \looseness=-1}
        \label{fig:joint-surprisal-variance}
\end{figure*}

\section{Experimental Setup}
\subsection{Data}
This work uses the publicly available Wiki40b dataset \cite{Guo_Dai_Vrandečić_Al-Rfou_2020}, a large text corpus derived from Wikipedia articles. 
We use subsets of the Wiki40b dataset in 10 languages: English, Russian, French, German, Hindi, Farsi, Vietnamese, Indonesian, Hungarian, and Turkish. The first six represent the Germanic, Slavic, Romance, Indo-Aryan, and Iranian sub-families of the Indo-European language family. The latter four belong to the Austroasiatic, Austronesian, Uralic, and Turkic language families, respectively.
Turkish, Hindi, and Farsi have basic SOV word order, while the other languages have SVO order with Hungarian being mixed \cite{wals-81}.
Languages were chosen based on the amount of available data in the Wiki40b dataset, their typological properties (covering a range of families, canonical word orders, and morphological complexity), and availability of automatic dependency parsing models.\looseness=-1

The datasets are subsampled to yield approximately 20M words in the training set of each language and approximately 1M words in the test and validation sets. 
We automatically generate dependency parses for all sentences using the UDPipe parser \cite{udpipe:2017}, yielding syntactic representations in the UD paradigm. 
We then apply each of the counterfactual orderings introduced in  \cref{sec:counterfactual-grammars} to the original data to create parallel corpora for each language.
Sentences are stripped of punctuation (as determined by the dependency parser's \textsc{Punct} label) and are lowercased. Periods are added back in to mark the end of sentences, regardless of what the original final punctuation was. 
Sub-word tokenization is then applied to the corpora using a byte-pair encoding (BPE) model, trained with a fixed vocabulary size of 30K tokens and using the algorithm of \citet{sennrich-etal-2016-neural}.
\footnote{All variants of the same language are tokenized using the same BPE model, trained on a sample of 100K documents from all variants; BPE tokens could not cross word boundaries for compatibility with different word orders.}
\looseness=-1

\subsection{Language Modeling}

For each variant of each language, we train a Transformer language model \cite{vaswani2017attention} using \texttt{fairseq} \cite{ott2019fairseq}. Models are trained on document-level inputs, with a maximum length of 512 tokens; this means that each token is predicted with the preceding material of the entire document as context. Each model is trained with early stopping, halting training after no improvement in validation loss for three epochs.
The Adam optimizer was used \cite{kingma2017adam}, with a learning rate of 0.0005,  weight decay of 0.01, and dropout of 0.1. Training scripts are available in the project's GitHub repository.$^{\ref{note1}}$ 
In all of our analyses, we use the  word-by-word surprisals estimated using our trained models on their corresponding held-out test sets. Note that we do not consider the designated $\textsc{eos}$ symbol in the computation of any of our UID-related metrics. 
In the case that a word is comprised of multiple sub-word tokens, we aggregate their surprisals by summation, since surprisal decomposes additively.

\section{Results} \label{sec:results}

Estimates of mean per-word surprisal on the test set are in \cref{fig:joint-surprisal-variance}A. Consistent with the results of \citet{Hahn_Jurafsky_Futrell_2020}, our trained models for nearly all counterfactual variants assign higher per-word surprisal to their respective test sets than the \variantReal{} models assign to theirs. Across all 10 languages, \variantReverse{} has mean surprisal close to, but consistently slightly higher than, that of the real ordering. \variantSortFreq{} and \variantSortFreqRev{} have mean surprisals close to or below those of \variantReal{}.\looseness=-1

\begin{figure*}
        \centering
        \includegraphics[width=0.9\linewidth]{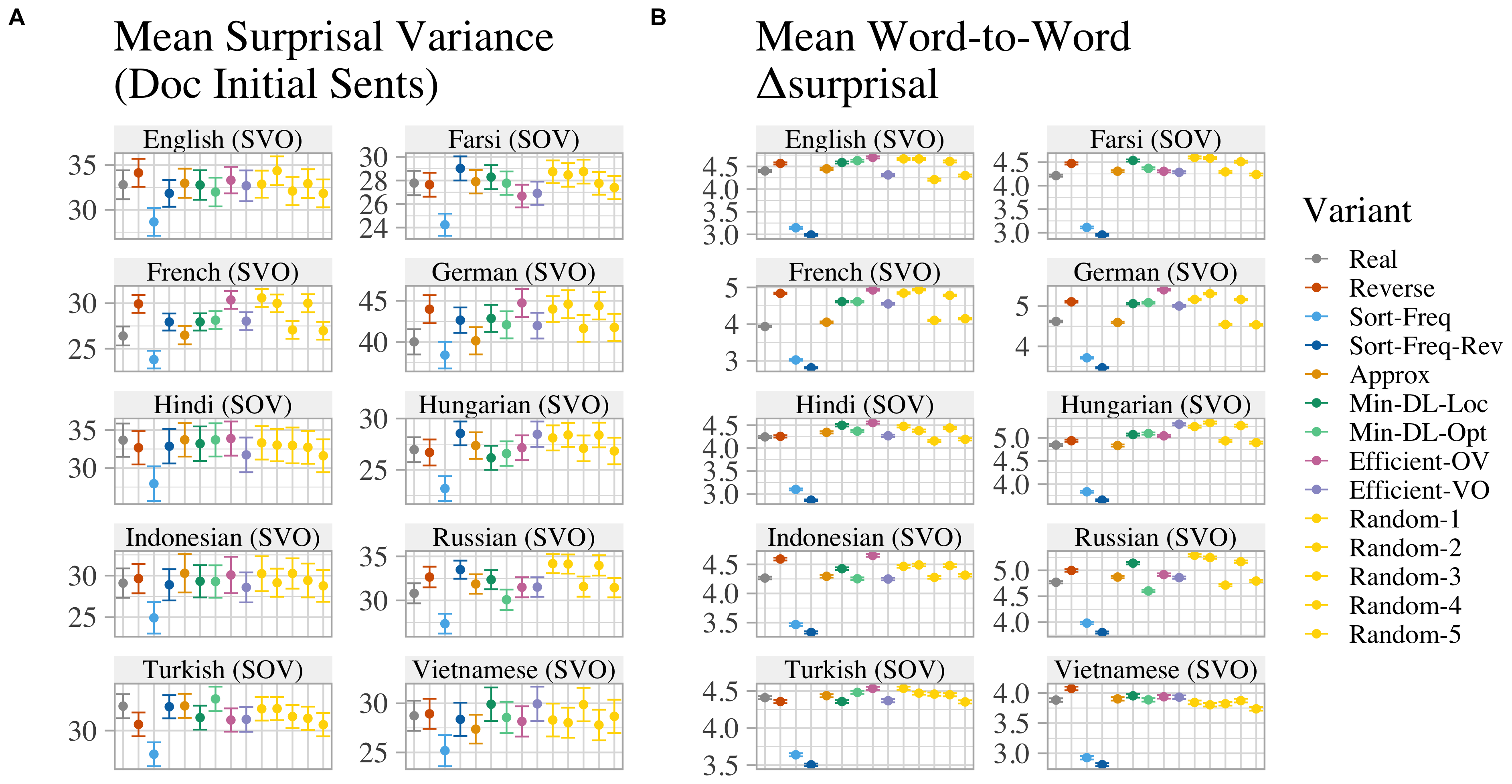}
        \caption{A. Surprisal variance ($\uidv$) for document-initial sentences only. B. Mean squared word-to-word change ($\uidl$) in surprisal. Error bars denote 95\% CI of the  mean.\looseness=-1}
        \label{fig:joint-doc-initial-uid-loc}
\end{figure*}

Estimates of mean surprisal variance ($\nuidv$) over sentences are shown in \cref{fig:joint-surprisal-variance}B. Notably, there is a dissociation between the rank order of variants according to mean surprisal and according to $\nuidv$: variants with similar mean surprisals did not necessarily have similar $\nuidv$ scores, and vice versa, suggesting that information uniformity and mean surprisal can vary independently of each other. Our main observations are as follows:
(i) In all languages except Turkish and Hindi, our estimates of $\nuidv$ for \variantReal{} are lower  than those for \variantReverse, despite the variants' similarities in mean surprisal. 
(ii) As predicted, the \variantSortFreq{} baseline has $\nuidv$ equal to or lower than that of \variantReal{}. 
(iii) The other counterfactual variants typically exhibit higher $\nuidv$ than \variantReal{}, with the exception of mixed results for \variantSortFreqRev{}.
(iv) The \variantVO{} variants typically have lower $\nuidv$ than \variantOV{} (with Hungarian being a noteworthy exception), which supports findings based on toy grammars showing that SVO orders are more uniform than SOV orders \citep{maurits2010why}.
Crucially, these results are qualitatively similar using the $\nuidl$ metric  (\cref{fig:joint-doc-initial-uid-loc}B).

\begin{figure*}[!ht]
        \centering
        \begin{tikzpicture}
            \node[anchor=south west,inner sep=0] (image) at (0,0){\includegraphics[width=1.0\linewidth]{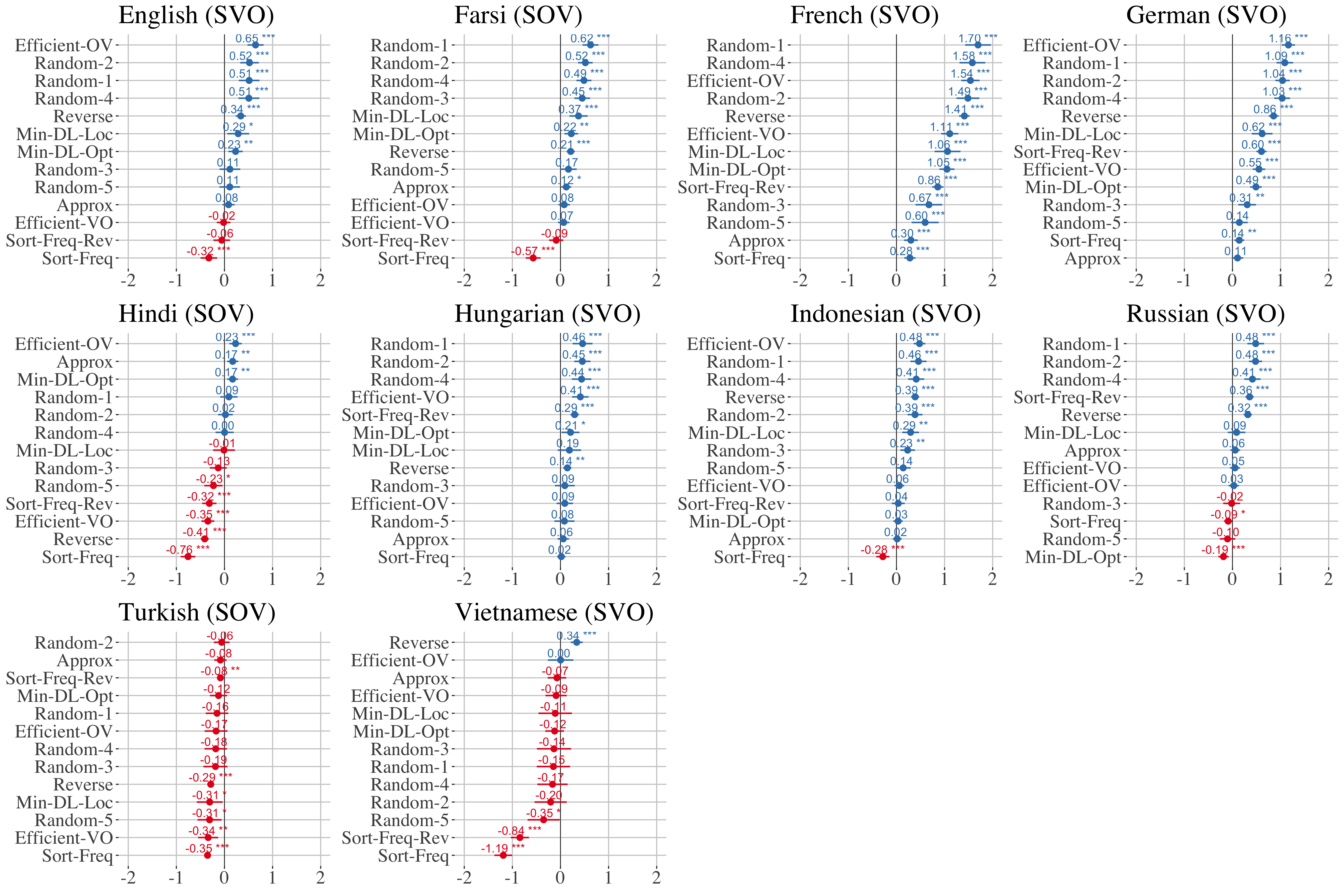}};
            \begin{scope}[x={(image.south east)},y={(image.north west)}]
                \draw (0.75, 0.14) node[text width=18em] {\small Figure \ref*{fig:uidp-regression}: Linear regression coefficient estimates when predicting $\nuidp$ as a function of mean surprisal, variant, and dataset size. The reference level for variant is \variantReal{}, so positive coefficients (blue) indicate variants with greater $\nuidp$, i.e., less uniformity, than real language.};
            \end{scope}
            
        \end{tikzpicture}
        \caption{}
        \label{fig:uidp-regression}
        \vspace{-5pt}
\end{figure*}

To fairly compare variants using the $\nuidp$ metric, we first need to account for the fact that, unlike surprisal variance, the metric is sensitive to shifts in mean surprisal. To control for this, we  fit a regression model predicting the $\nuidp$ score based on three variables: the mean surprisal, the grammar variant, and the dataset size (20M, 6.6M, and 3.3M words). We train multiple language models for each language-variant combination (3 dataset sizes and 2 random seeds), resulting in 84 data points per language. We apply treatment coding to the variants, with \variantReal{} as the reference level. \cref{fig:uidp-regression} shows the resulting estimates of the coefficients for each variant, where a coefficient should be positive if that variant is less uniform than \variantReal{}. Qualitatively, the regression results match the results given by $\nuidv$ and $\nuidl$: \variantReal{} is more uniform than \variantReverse{} in SOV languages, \variantSortFreq{} is the only counterfactual variant that is consistently more uniform than \variantReal{}, and \variantVO{} is more uniform than \variantOV{} in most languages; the opposite is true in Hungarian and the difference is negligible in Russian.

\captionsetup{labelformat=empty}
\captionsetup{labelformat=original}

\section{Discussion}
We offer a discussion of the results observed in \cref{sec:results}, including their implications for the role of functional pressures in language.

\subsection{Differences in mean surprisal}
Across 10 typologically diverse languages, we find that Transformer LMs learn to predict data from real word orders better than data from counterfactual orders, with the exception of the \variantSortFreq{} and \variantSortFreqRev{} variants. This suggests that these LMs' inductive biases somehow favor properties of real languages, in line with previous work on other modeling architectures \citep{Gildea_Jaeger_2015, Ravfogel_Goldberg_Linzen_2019}. This is not surprising, given that commonly used architectures and hyperparameters have been selected specifically based on their good performance on real language tasks. 
Unlike in $n$-gram models, the precise inductive bias of Transformer models that favors real word orders is not transparent and merits further study.\footnote{Notably, \citet{white-cotterell-2021-examining} show that there is a large variation in how Transformer language models perform in toy languages with diverse word orders; they, however, do not find evidence that Transformers perform better on the most frequently occurring orders (as opposed to, e.g., OVS and VOS word orders, which are found in few languages).}\looseness=-1

\subsection{Differences between \variantReal{} and \variantApprox{}}
We observe that despite the similarities between the \variantReal{} and \variantApprox{} variants of a given language, the latter are consistently assigned higher mean surprisal by their respective LMs. Meanwhile, the various UID metrics show similar results for \variantReal{} and \variantApprox{}, suggesting that the greater flexibility of \variantReal{} is not responsible for UID differences in our results. This is somewhat surprising, since it may appear that such flexibility is what enables speakers' choices, which have been previously discussed as contributing to UID. However, many speaker choices that potentially impact UID, such as word choice, active versus passive voice, and optional words, are not captured by this difference in flexibility between \variantReal{} and \variantApprox{}.\looseness=-1 

\subsection{Greater uniformity of \variantReal{} over \variantReverse{} in SVO languages}
While mean surprisal is always very close for \variantReal{} and \variantReverse{} grammars, \variantReverse{} is less uniform in 8 out of 10 languages, including all SVO languages. This held across multiple operationalizations of UID, with the exception of mixed results for Hungarian, a language with considerable flexibility in word order. Thus, while both \variantReal{} and \variantReverse{} orders are learned approximately equally well by language models, they differ in how uniformly they distribute information.

One key difference between \variantReal{} and \variantReverse{} is that insofar as \variantReal{} sentences exhibit a tendency to mention entities from the end of a given sentence close to the beginning of the next one, \variantReverse{} does not preserve this property. 
For example, the pair of sentences \texttt{``I like dogs. They are friendly.''} would become \texttt{``Dogs like I. Friendly are they.''}; note that the distance between antecedent and pronoun is significantly increased. 
This feature of the \variantReverse{} raises the possibility that the uniformity patterns we observe are due to speaker choices taking cross-sentence dependencies into consideration. 
To minimize the influence of cross-sentence dependencies, we can consider only sentences occurring at the start of a document, which cannot refer to previous sentences. \cref{fig:joint-doc-initial-uid-loc}A shows that the tendency for \variantReal{} to have lower surprisal variance than \variantReverse{} still holds in this setting across most languages. This suggests that cross-sentence dependencies alone cannot fully explain the observed differences in information uniformity.\looseness=-1

Notably, our results show that the UID preference for \variantReal{} over \variantReverse{} is not consistently present in languages with basic SOV order (Turkish, Hindi, and Farsi). We propose the following explanation for this result: As argued in \citet{maurits2010why}, SVO languages tend to have more uniform information density profiles than SOV languages -- a finding supported by our empirical results in which \variantVO{} had lower surprisal variance than \variantOV{} in 9 out of 10 languages. Unlike the short, simple sentences of \citeauthor{maurits2010why}, however, the present study considers long and complex sentences where speaker choices have considerable opportunity to influence information uniformity, in addition to the role of basic word order. These choices include whether to use a pronoun, whether to use an active or passive construction, and what order to present a conjunction or list of items, among others. 
Importantly, speakers make choices conditional on the forward ordering of real language, so we expect that the choices made in an attempt to increase UID -- which constitutes a non-trivial percentage of utterances \citep{levy2007speakers} -- would have a  greater effect on UID in \variantReal{} than in \variantReverse{}. In SVO languages, the effects upon UID of basic word order and speaker choices both go in the same direction -- towards more uniformity. In SOV languages, these effects conflict -- the basic word order is non-optimal in terms of UID, and so uniformity can theoretically be increased by a transformation to \variantReverse{}, while speaker choices are presumably already mostly optimal in \variantReal{}. This may explain the heterogeneous patterning among the three SOV languages. 

Furthermore, these results can potentially shed light on an important question in linguistic typology: Why are some basic word orders more common than others? According to some theories, SOV order (the most typologically common) is the most natural for expressing events with subjects and objects \citep{Goldin-Meadow_So_Özyürek_Mylander_2008, Gibson_Piantadosi_Brink_Bergen_Lim_Saxe_2013, futrell2015crosslinguistic}. If these theories are correct, an evolutionary pressure on languages to shift from SOV to SVO could help account for the prevalence of SVO languages, which are nearly as common as SOV ones. A pressure for information uniformity offers one such account.

Finally, \citet{pimentel-etal-2021-disambiguatory} has recently shown that the distribution of per-phone information \emph{within words} is more uniform when analysed in reverse order than in forward order -- the opposite of what we observe on our sentence-level analysis.
This difference may suggest qualitatively distinct information-theoretic pressures being present at the lexical and sentential levels and is a potential topic for further study.

\subsection{Other Variants}

The variants designed to minimize dependency length, \variantDlLoc{} and \variantDlOpt{}, showed mixed results in terms of information uniformity compared to \variantReal{}. The random grammars fell into two groups: \variantRandom{1}, \variantRandom{2}, and \variantRandom{4} tended to be less uniform than \variantReal{}, while \variantRandom{3} and \variantRandom{5} tended to be similar in uniformity to \variantReal{}. Since random grammars have fixed but uncorrelated directions of syntactic relations, these cross-linguistically consistent patterns suggest that some settings of the parameterized grammar are inherently more favorable from the perspective of UID than others.

The only counterfactual word order to consistently have a higher degree of information uniformity than the real orders was the highly constrained \variantSortFreq, which turns sentences into sorted word lists. Thus, while it appears possible to improve on real word orders' information uniformity, this comes at the cost of massive syntactic ambiguity and reduced expressivity.\looseness=-1

\subsection{Robustness to Dataset Choice} \label{sec:dataset-choice}
\begin{figure}
        \centering
    \includegraphics[width=\linewidth]{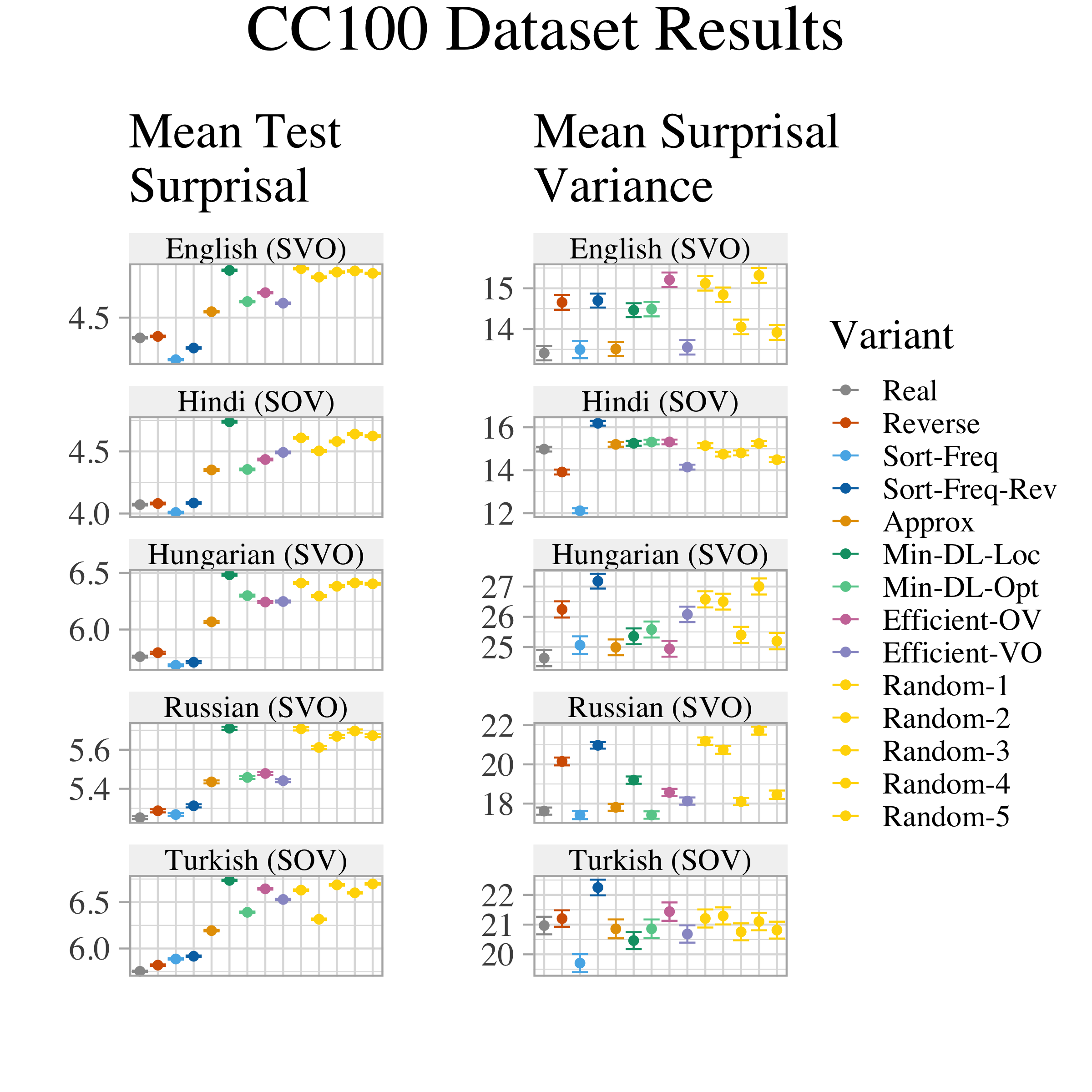}
    \vspace{-30pt}
        \caption{Surprisal mean and variance for a subset of languages on the CC100 dataset. Error bars denote 95\% CI.\looseness=-1}
        \label{fig:surprisal-variance-cc100}
\end{figure}

In this study, the chosen dataset (Wiki40b) contains formal writing that may not exhibit the same communicative pressures as spoken language. It is largely devoid of first and second person pronouns, interrogatives, and other features common in everyday speech; further, it may have disproportionate amounts of translationese \cite{Koppel_Ordan_2011}.
As a supplementary analysis, we repeated the experiments on the CC100 dataset \citep{Conneau_Khandelwal_Goyal_Chaudhary_Wenzek_Guzmán_Grave_Ott_Zettlemoyer_Stoyanov_2020}, using only a subset of languages due to computational constraints. This dataset is sourced from a web crawl and therefore contains a wider range of genres and styles than Wiki40b. $\nuidv$ scores for these experiments are shown in \cref{fig:surprisal-variance-cc100}. The results qualitatively match the patterns from the Wiki40b experiments in the following ways: (i) better $\nuidv$ scores for \variantReal{} than for \variantReverse{} among SVO languages, (ii) better $\nuidv$ scores for \variantVO{} than \variantOV{} in most languages (with Hungarian again being an exception), and (iii) the only variant that has higher uniformity that \variantReal{} across a majority of languages is \variantSortFreq{}.

\section{Conclusion}

In conclusion, we have empirically demonstrated that in many languages, real word orders distribute information more uniformly than a range of counterfactual orders. The fact that this pattern holds in every SVO languages but is mixed
among SOV languages lends support to the view that SVO basic word order is preferable to SOV order from the perspective of maximizing UID. We posit that there are two potential sources of optimization within a language for greater UID: language evolution favoring word orders that produce less variance in information content, and speaker choices in favor of constructions that smooth the information profile of utterances. Our results are consistent with the UID hypothesis, and support the idea that communicative pressures (operationalized in terms of information theory) influence the structure of human language.

\section*{Acknowledgements}
We thank our action editor, Mark-Jan Nederhof, and the anonymous reviewers for their detailed feedback on this paper. 
CM was supported by the Google PhD Fellowship. 
TP was supported by a Facebook PhD Fellowship. 
This work was supported by NSF grant BCS-2121074 to RPL.

\bibliography{tacl2021}
\bibliographystyle{acl_natbib}

\end{document}